\documentclass{article}
\usepackage{amsmath,amsfonts}
\usepackage{algorithmic}
\usepackage{algorithm}
\usepackage{array}
\usepackage[caption=false,font=normalsize,labelfont=sf,textfont=sf]{subfig}
\usepackage{textcomp}
\usepackage{stfloats}
\usepackage{url}
\usepackage{verbatim}
\usepackage{graphicx}
\usepackage{authblk}
\usepackage{cite}
\usepackage[colorlinks,allcolors=blue]{hyperref}
\usepackage{lineno,hyperref}

\newtheorem{defn}{Definition}[section]

\title{Can persistent homology whiten Transformer-based black-box models? A case study on BERT compression}
\author[1]{Luis Balderas
	\thanks{Corresponding author. luisbalru@decsai.ugr.es}}

\author[2]{Miguel Lastra}

\author[1]{Jose M. Benítez}

\affil[1]{Department of Computer Science and Artificial Intelligence, DiCITS, iMUDS, DaSCI, E.T.S.I.I.T. University of Granada, Spain}
\affil[2]{Department of Software Engineering, DiCITS, iMUDS, DaSCI, E.T.S.I.I.T. University of Granada, Spain}

\begin{document}

\maketitle

\begin{abstract}
Large Language Models (LLMs) like BERT have gained significant prominence due to their remarkable performance in various natural language processing tasks. However, they come with substantial computational and memory costs. Additionally, they are essentially black-box models, challenging to explain and interpret. In this article, we propose Optimus BERT Compression and Explainability (OBCE), a methodology to bring explainability to BERT models using persistent homology, aiming to measure the importance of each neuron by studying the topological characteristics of their outputs. As a result, we can compress BERT significantly by reducing the number of parameters (58.47\% of the original parameters for BERT Base, 52.3\% for BERT Large). We evaluated our methodology on the standard GLUE Benchmark, comparing the results with state-of-the-art techniques and achieving outstanding results. Consequently, our methodology can ``whiten'' BERT models by providing explainability to its neurons and reducing the model's size, making it more suitable for deployment on resource-constrained devices. 
\end{abstract}

\section{Introduction}
Over the last years, Large Language Models (LLM) like BERT (Bidirectional Encoder Representations from Transformers) \cite{devlin2019bert} have achieved phenomenal results in a wide variety of natural language processing (NLP) tasks. These models, pre-trained on large amounts of data, are available for use by the scientific community, either as a semantic analytical tool or to build fine-tuned solutions for specific problems.

However, these models suffer from two weaknesses. First, they are computationally and memory-intensive, making it challenging to deploy them on devices with limited resources. Second, they are black-box models: due to the large number of neurons, layers, parameters, and data transformations (e.g., in Attention layers \cite{vaswani2023attention}), it is practically impossible to explain the internal state of the network at any given moment and provide interpretability to the final output.

In this article, we propose Optimus BERT Compression and Explainability (OBCE), a methodology based on homology theory to provide explainability to the BERT model. We have chosen BERT as our reference model given the fact that is one of the foundational models that laid the groundwork for LLM encoders. Several subsequent LLMs have been developed based on the principles introduced by BERT, such as RoBERTa \cite{roberta}, DistilBERT \cite{distilbert}, ALBERT \cite{albert} or TinyBERT \cite{tinybert}. Following the definitions of explainability outlined in \cite{8631448}, we focus on the internal representation of the neural network, considering the individual role of each neuron when the model makes an inference. After selecting a dataset, we analyze the outputs of each neuron on that dataset and use zero-dimensional persistent homology, extracting the topological characteristics of these outputs and providing an assessment of the neuron's importance within the network. This way, we can identify neurons that contribute more information to the overall computation of the network, allowing us to remove those that contribute less. As a consequence, our methodology becomes an effective method for compressing BERT.

To the best of our knowledge, this is the first approach that uses persistent homology to explain and compress BERT. To measure the effectiveness of our methodology, we have designed an extensive experimentation based on the natural language processing tasks proposed in GLUE (General Language Understanding Evaluation) benchmark \cite{wang-etal-2018-glue}, achieving results that outperforms other state-of-the-art techniques for BERT compression. Our contributions can be summarized as follows:

\begin{itemize}
	\item We have developed a methodology for providing explainability to LLMs derived from BERT, in terms of the individual role of each neuron when the model makes an inference, based on zero-dimensional persistent homology, capable of analyzing the topological features of the outputs of each neuron.
	\item We applied this methodology to two versions of the BERT model, interpreting the topological characteristics as a tool to assess the importance of neurons, simplifying those that contribute less information, and generating a compressed version of the BERT model.
	\item We evaluated the performance of the simplified models using the GLUE Benchmark and compared the results with other state-of-the-art compression techniques, demonstrating the effectiveness of our methodology for model explicability and compression.
\end{itemize}

The rest of this paper is structured as follows: In Section 2, we introduce the state-of-art of different approaches using persistent homology and deep learning models to solve problems and BERT compression techniques. In Section 3, we describe our proposal. In Section 4 our methodology is experimentally analyzed and discussed. Section 5 highlights the conclusions. 

\section{Previous work}

In this section, we present the most relevant articles from the state of the art related to our work. To the best of our knowledge, there is no technique in the literature that uses persistent homology to prune neural networks. Therefore, we first present techniques that utilize homology theory and deep learning to solve scientific problems. Finally, we present pruning techniques for LLMs, which will serve as a reference to validate our methodology.

\subsection{Persistent homology applied to machine learning problems}

In recent years, numerous machine learning methods based on persistent homology, which is a mathematical method used in topological data analysis to study features of data, have been proposed. Some of them involve feature extraction through the analysis of persistence diagrams (Birth-Death diagrams) and persistence barcodes. Additionally, there are methods based on kernel with persistent homology, which can establish similarity metrics between barcodes \cite{Mileyko_2011}. From a mathematical perspective, solutions based on persistent homology have been presented to regularize the internal representation of deep neural networks applied to image classification \cite{WOS:000940775900001}, \cite{WOS:000775955400001}.

Persistent homology combined with machine learning also finds applications in the fields of biology and chemistry, particularly in protein analysis. Deep neural networks are not directly applicable to molecular data, as they are characterized by complex three-dimensional structures. Therefore, the use of topological representations and features is essential for solving protein classification problems \cite{Pun2022}. They are also highly useful for identifying structures and functions in a protein sequence \cite{WOS:000848431900001}, for automatic protein annotation \cite{WOS:000472576800003}, or in chemistry, for tasks like the simultaneous prediction of partition coefficients and aqueous solubility \cite{WOS:000436939500008}.

Getting closer to our topic, homology theory has been used to analyze natural language. Specifically, in \cite{doi:10.1080/17538947.2023.2239794}, TopoBERT is presented as a toponym recognition module based on one-dimensional CNNs and BERT.

\subsection{Brief description of the BERT model}

BERT, an acronym for "Bidirectional Encoder Representations from Transformers," \cite{devlin2019bert} is a language model developed by Google that has revolutionized natural language processing. Its primary innovation lies in its ability to comprehend a word's context within a text by analyzing all surrounding words, both to the left and right. Unlike previous language models, which were unidirectional and predicted words based on previous ones, BERT takes the entire context into account. 

\begin{figure*}[!t]
	\centering
	\includegraphics[width=\textwidth]{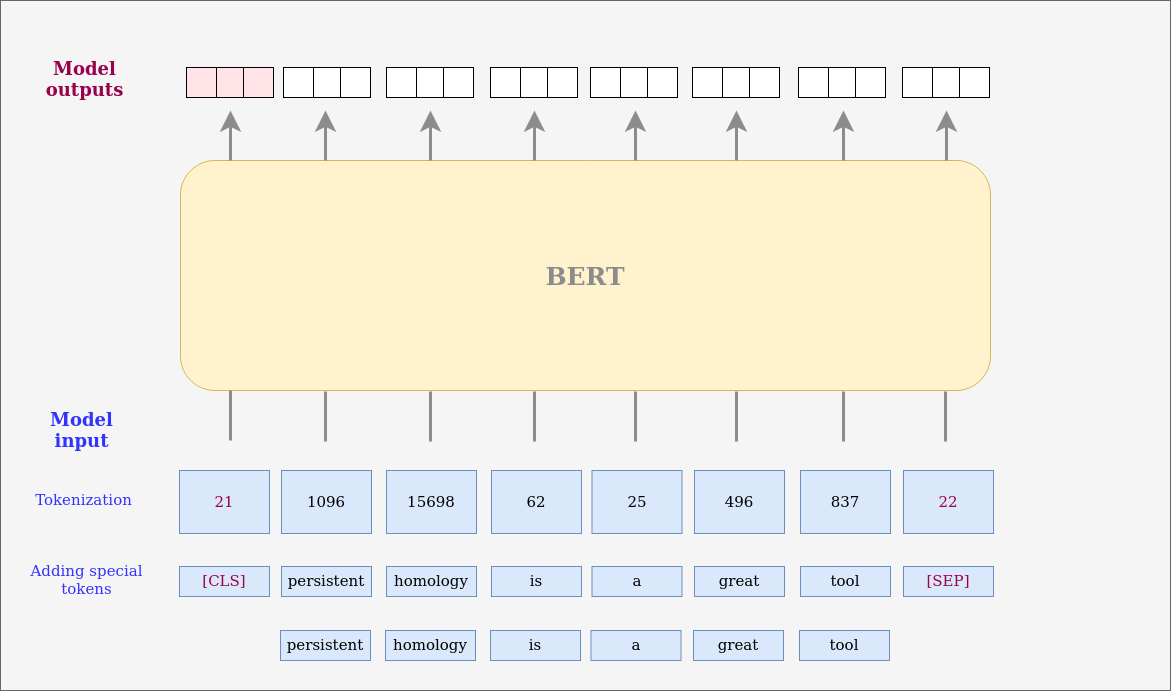}
	\caption{Usage of BERT involves taking a sentence, adding the special tokens [CLS] and [SEP], tokenizing the words, and using these tokens as input for the neural network.}
	\label{bert}
\end{figure*}

BERT is built upon the Transformer architecture, a deep neural network autoencoder. The Transformer architecture employs multi-head attention to process sequences. Before the network processes the information, the texts are tokenized, meaning texts are divided into a set of tokens that represent the fundamental semantic content of the word and are translated into their numerical position in the model's vocabulary. Additionally, special tokens are added, such as [CLS], marking the start of a sentence; [SEP], marking the end, and [PAD], signifying the end of padding. Some research papers from the literature, such as \cite{clark2019does}, show that the [CLS] token, besides marking the beginning of a sequence, provides an aggregated representation of the input. In fact, it is used in text classification tasks and sentiment analysis. Figure \ref{bert} shows an example of how sequences are generated to be a valid input for BERT model.

To train BERT, an extensive amount of web text is used. During the training process, BERT learns to predict hidden words in complete sentences, enabling it to better understand how words relate to each other in a given context. 

This pre-trained model can be fine-tuned for specific tasks such as text classification, entity tagging, machine translation, and more. Due to its context-awareness, BERT has outperformed many records in various natural language processing tasks and has become an essential tool in this field.

\begin{figure*}[!t]
	\centering
	\includegraphics[width=0.5\textwidth]{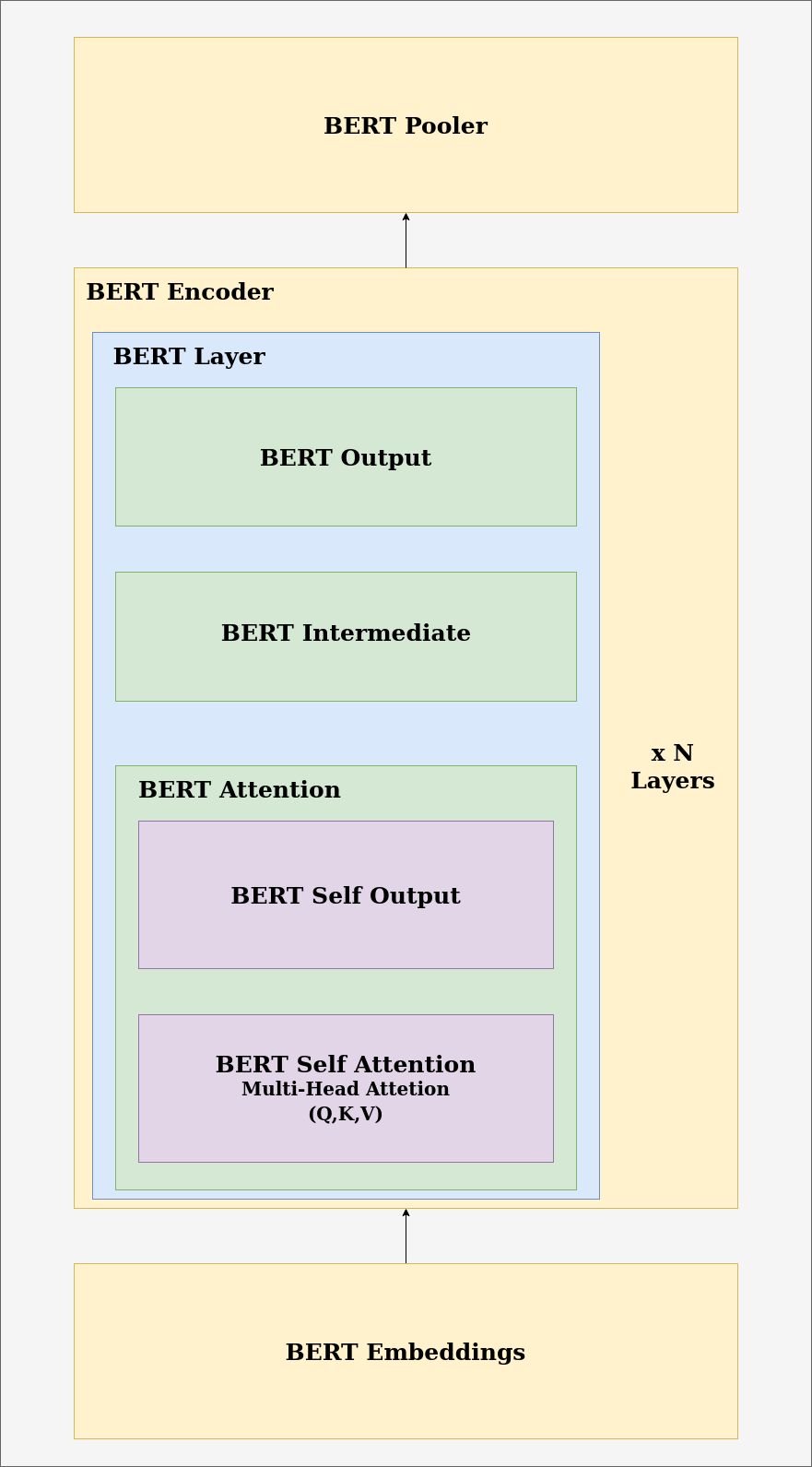}
	\caption{Representation of the BERT architecture. It is composed of an embedding module, followed by the Encoder part, which consists of $N$ BERT Layers (12 or 24, depending on whether it's BERT Base or Large). Within it, three main components stand out: the Attention layer, the Intermediate layer, and the Output layer. After the Encoder, BERT has a Pooler layer.}
	\label{bert-detalles}
\end{figure*}

Each BERT Layer, as it is part of the Encoder part of the Transformers architecture, consists of three components: the Multi-Head Attention layer, comprised of the Q (query), K (key), V (value) matrices, and Attention Output; the intermediate layer; and the output layer. Figure \ref{bert-detalles} shows the complete representation of the BERT architecture. Besides, more detailed information can be found in \cite{devlin2019bert} and \cite{vaswani2023attention}.

Let $L$ be the number of layers, $H$ the hidden size, and $A$ the number of self-attention layers of the model, there are two implementations of the BERT model: the BERT Base Cased model \cite{bertbasecased} ($L=12$, $H=768$, $A=12$, total number of parameters $= 110M$) and the BERT Large Cased model \cite{bertlargecased} ($L=24$, $H=1024$, $A=16$, total number of parameters $= 340M$). 

\subsection{BERT model pruning methods}

As we mentioned earlier, to the best of our knowledge, there are no pruning methods for LLMs based on persistent homology. However, there are compression algorithms for the BERT architecture that provide competitive results. These methods can be divided into structured and unstructured approaches \cite{ganesh-etal-2021-compressing}.

Among structured methods, some of them focus on pruning attention heads exclusively. In \cite{Lee2021-bo} a regularization method for BERT is proposed using reinforcement learning to prune attention heads. In \cite{michel2019sixteen}, a forward and backward pass is used to calculate gradients, which are used as an importance score for each attention layer. \cite{voita-etal-2019-analyzing} suggests constructing a loss function that minimizes both classification error and the number of used heads, pruning unproductive ones while maintaining performance. \cite{WOS:000890309300033} presents a technique with three stages: in the first, $n$ pruning strategies are generated with the same pruning ratio. In stage II, $n$ candidates from the training set are evaluated, and after all iterations, the one that performs the best on a subnetwork is chosen as the best candidate. This best candidate undergoes fine-tuning to obtain a good subnetwork.

Among unstructured methods, \cite{chen2020lottery} uses unstructured magnitude pruning to find subnetworks with sparsity levels between 40\% and 90\%. It concludes that those with 70\% sparsity, using the masked language modeling task, are universal and can solve other tasks without losing accuracy. \cite{guo2019reweighted} proposes a weight pruning algorithm that integrates reweighted L1 minimization with a proximal algorithm. You can find more information on pruning algorithms for LLMs in \cite{WOS:000804984600001}.

Quantization algorithms have also been proposed, such as \cite{shen2019qbert}, which uses second-order Hessian matrix information for quantizing BERT models to low precision. Hardware-based techniques have been proposed too \cite{li-etal-2020-efficient-transformer}. Finally, there are knowledge distillation-based algorithms, like the one presented in \cite{WOS:001041973100003}, which is based on parameter retention and feed forward network parameter distillation.

\section{Our proposal}

In this article, we propose Optimus BERT Compression and Explainability (OBCE), a novel technique to endow explainability to BERT-based models, in terms of the role of individual units in the inference and learning process, through the application of homology theory and, at the same time, allowing to derive a simplified yet effective version of the model. Specifically, using the BERT architecture as a reference, we employ persistent homology as a fundamental tool for analyzing the topological characteristics of layer outputs when evaluating the network on a predetermined dataset, drawing conclusions to discard neurons that do not make a relevant contribution to the model. In this section, we first provide an intuitive geometric description of using zero-dimensional persistent homology to the vectors generated by hidden layers of a neural network. Finally, we present the proposed methodology for introducing explainability to the model and simplifying those less relevant units.

\subsection{An intuitive geometric description of persistent homology applied to LLM explanations}

Homology theory is a branch of algebraic topology increasingly used in data science. In particular, persistent homology is a powerful tool for studying patterns in data. Its mathematical foundation, available in Appendix I, is deep and complex with a significant algebraic burden. However, it can be given a much more intuitive geometric interpretation, which we will explain below.

\begin{figure*}[!t]
	\centering
	\includegraphics[width=\textwidth]{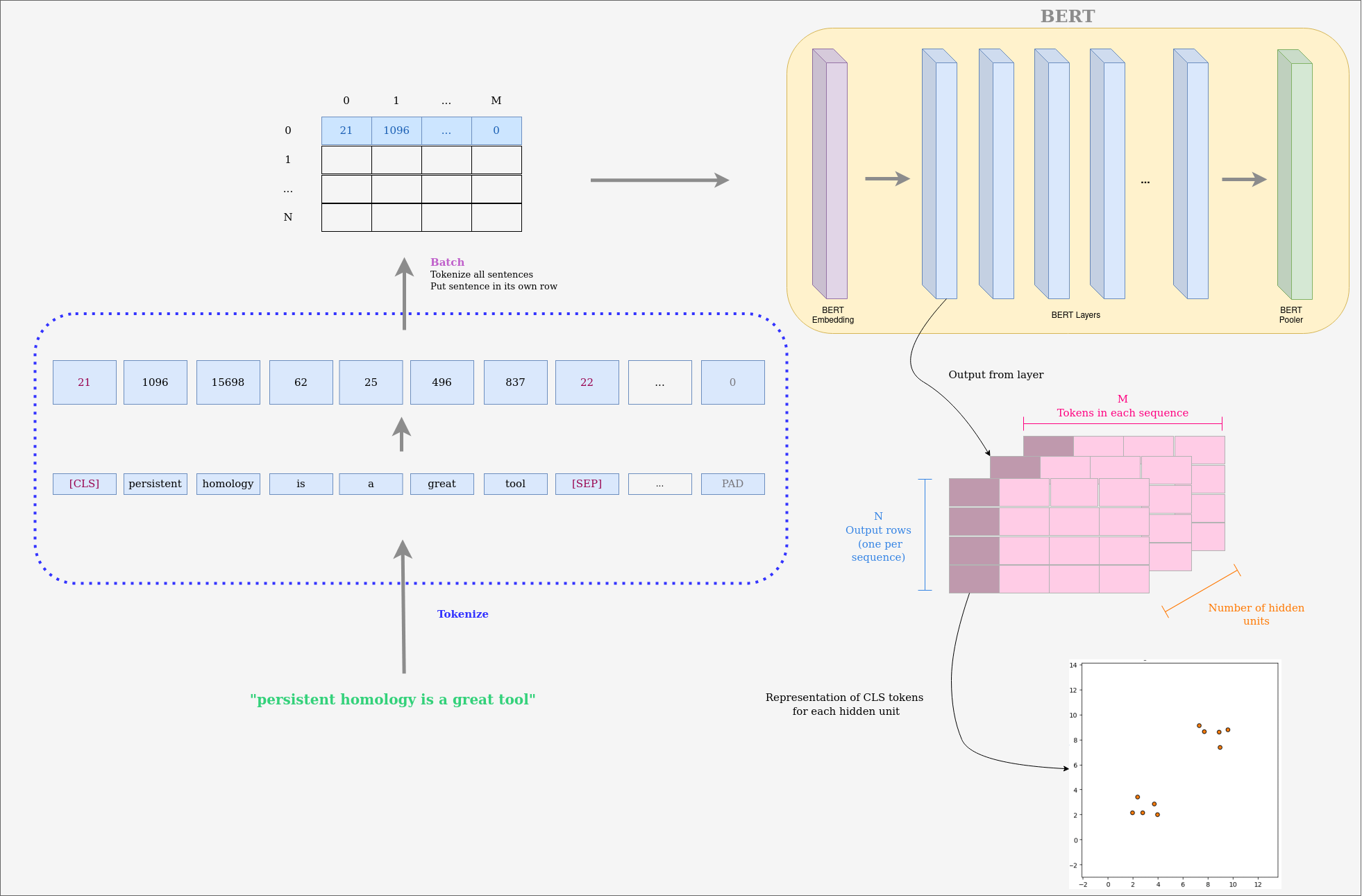}
	\caption{Representation of information through the BERT model and subsequent extraction of values from intermediate dense layers. The process begins with the processing of a set of sentences. These sentences are tokenized, adding special tokens such as [CLS], which marks the beginning of a sentence, or [SEP], which marks the end. The neural network's input must have a fixed token size per instance. Since sentences can vary in length, the maximum length is found and taken as a reference, filling the remaining gaps with 0 and the [PAD] token. Once the input of $N$ sentences with length $M$ is constructed, it is fed into the neural network. In the lower right part of the image, the output of any dense layer is represented. The output is a three-dimensional matrix: the number of sentences in the input ($N$), the length of those tokenized sentences ($M$), and the number of hidden neurons in that layer. Since the [CLS] token (first column of each matrix) usually contains information that encapsulates the semantics of the entire sentence, in this methodology, we use the values of the [CLS] token for subsequent analysis. For the sake of clarity, in this example, we show the representation of the [CLS] token in two dimensions}
	\label{diagrama1}
\end{figure*}

In Figure \ref{diagrama1}, we find a complete example of data processing in BERT. Starting with a set of sentences, after being tokenized, they are fed into the network. The output of any neuron in a hidden layer, which serves as input for the next layers, consists of vectors that, in an abstract sense, could be represented as points in a hyperspace. The geometric distribution of these points can provide us with meaningful information about the behavior and role that the corresponding neuron plays within the data flow of the network. In Figure \ref{diagrama1}, as an example of the output of a specific neuron, it is very clear that there are two groups of points. However, conducting this type of analysis in high-dimensional spaces is complex. Zero-dimensional persistent homology helps us explain the role of each neuron.

\begin{figure*}[!t]
	\centering
	\includegraphics[width=0.75\textwidth]{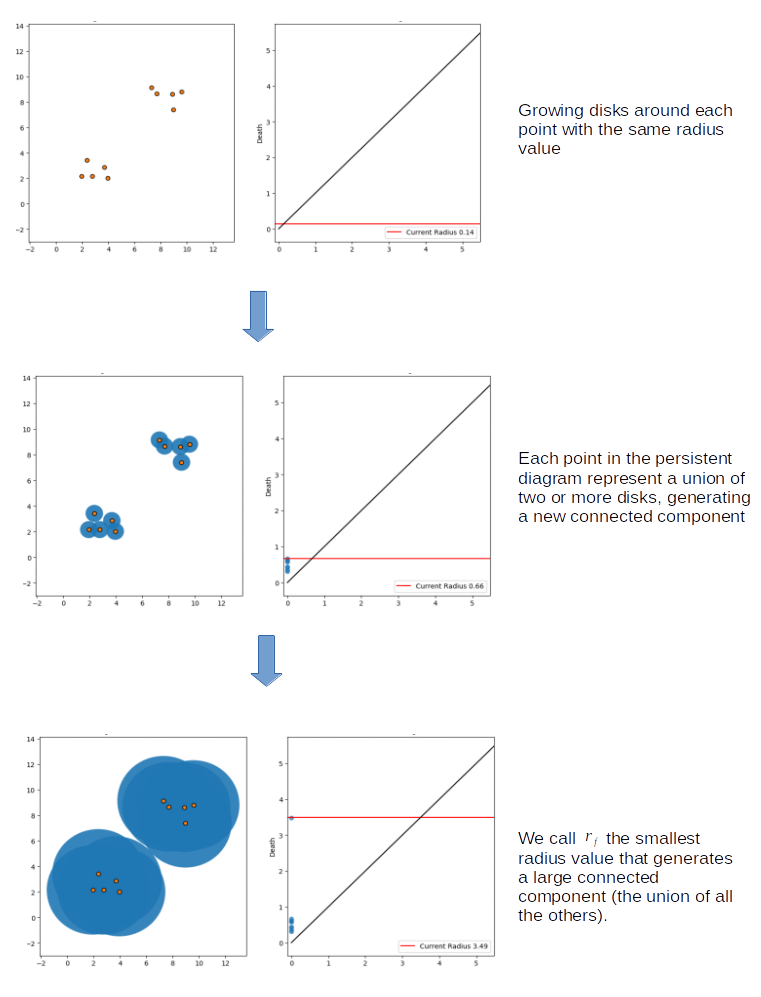}
	\caption{Application of persistent homology on the output of a neuron at three specific moments. On the left, you can see that each of the points comprising the output becomes the center of a disk whose radius grows uniformly for all points. On the right, we represent the Birth-Death diagram for persistent homology of dimension zero. Each blue point corresponds to the disappearance of a connected component after collapsing with another. The last moment depicted in the figure represents the point at which the value of $r$ is reached for which all connected components first merge. We call this value $r_f$, and it is crucial in our methodology because it provides information about the importance of neurons based on their output within the neural network's data flow.}
	\label{diagrama2}
\end{figure*}

Once a neuron is selected, and its outputs are generated as vectors, we proceed to use persistent homology. Specifically, in Figure \ref{diagrama2}, you can see two graphs at three different time moments. The graph on the left represents the vectors generated, which are going to be converted into the centers of disks. These disks will evolve, growing uniformly with a common radius. This radius is represented on the graphics: on the left, called Birth-Death diagram, as the size of disks, and its value marked by the red line on the right. Notice that this red line will move upwards on the Y-axis of the graph on the right. Therefore, the Birth-Death diagram represents the evolution of connected components present in the graph on the right.

Since we are focusing on zero-dimensional persistent homology, all connected components (the vectors) start at time zero in the Birth-Death diagram. As the radius of the disks grows, the connected components will merge, ceasing to be independent and becoming larger connected components. When two connected components touch, they merge becoming a single area and a blue point is marked on the Birth-Death plot at the radius value that led to the merger. Remarkably, the radius grows until the last two connected components touch, creating a large single connected component that includes all the others. We will call $r_f$  the radius value at which all connected components collapse. Please, note that this is the smallest value for the radius at which all the connected componets are glued together. It will be fundamental in our analysis because it provides information about the distribution of output vectors from each neuron and the ``distance'' between them. This helps us assess whether the neuron has very uniform outputs (providing little information) or exhibit variability in its outputs (providing more information). All in all, this procedure allows to ``measure'' the variability in the output distribution of a neuron. It can be described through a simple geometric procedure. Nevertheless, it has well founded mathematical background.

\subsection{Explaining LLMs: BERT compression using persistent homology}

The explainability of a machine learning model is associated with its internal logic and the transformations that occur within it. The more explainable a model is, the greater the level of understanding a human can achieve in terms of the internal processes that take place when the model makes decisions \cite{e23010018}.

In this article, we designed a methodology focused on the explainability of LLMs, specifically the explainability of the BERT model. Based on the definitions proposed in \cite{8631448}, our methodology tries to explain the internal representation of deep neural networks, taking into account the role of individual units in the inference and learning processes. Concretely, we use zero-dimensional persistent homology to analyze the topological characteristics of neuron outputs for each layer, aiming to identify which of them play a more significant role in the information flow and, consequently, in the model's decision-making process. This allows for the removal of neurons that provide less information, compressing the network and making it more efficient. Algorithm \ref{alg:alg1} outlines our proposed methodology. In summary, our methodology starts by selecting a dataset for model evaluation. Next, we use zero-dimensional persistent homology as a tool to measure the importance of neurons. Then, using the persistence diagram, we analyze which neurons can be removed. Finally, we construct the simplified model and evaluate it on the GLUE Benchmark.

Next, we provide a detailed description of each of the steps outlined in Algorithm \ref{alg:alg1}, which constitute the methodology followed for the explainability and compression of the BERT models.

\begin{algorithm}[h]
	\caption{BERT explainability and compression through zero-dimensional persistent homology}\label{alg:alg1}
	\begin{algorithmic}
		\STATE 
		\STATE 1) Select the dataset to carry out the analysis.
		\STATE 2) Use the zero-dimensional persistent homology and the persistence diagram for the outputs of each unit and layer evaluated on the selected dataset.
		\STATE 3) Analyze the distribution of $r_f$ from the persistence diagram. Decide which units can be removed. 
		\STATE 4) Construct the simplified model.
		\STATE 5) Evaluate the simplified model with the GLUE Benchmark.
	\end{algorithmic}
	\label{alg1}
\end{algorithm}

\subsubsection{Dataset selection}

Any methodology aiming to work with deep neural networks requires a dataset to serve as a reference for conducting the necessary analyses. Since this article focuses on LLMs, specifically BERT, the selected dataset is textual. We chose the English Wikipedia \cite{wikipedia}, consisting of more than 20GB of sanitized text, including all entries from the Wikipedia in the mentioned language. To ensure that the texts, after tokenization, adhere to the input size constraint imposed by each neural network, we split the entries by periods, generating ``short texts'' that would be suitable for the neural network input. Finally, a random subset of these short texts is selected. Different dataset sizes have been tested, specifically 150, 700, 1500, and 3000 texts, yielding similar results in terms of the neurons that can be removed. In consequence, it was decided to fix the number of texts in the dataset at 150 to accelerate the runtime of the process.

\subsubsection{Using persistent homology to explain BERT Layer outputs}

Once the dataset has been selected for analysis, we proceed to use persistent homology to collect and analyze the topological characteristics expressed by each neuron in the network. After tokenizing the texts from the dataset, for each unit, we evaluate the dataset and extract the [CLS] tokens, given its importance in the semantic analysis of the input, as it was previously mentioned. Using these tokens, we construct the connected components which are the neuron outputs. After constructing the complex, the next step is to compute the connected components within it. These connected components or disks represent the clusters or groups of data points that are topologically connected in some way, and their persistence helps reveal meaningful structures in the data that persist over various scales. This information is crucial for understanding the topological features present in the neural network's outputs. The connected components evolve as the radius $r$ grows, until all connected components collapse into one. The  least value of $r$ which provokes the union of all the connected components, denoted as $r_f$, is crucial in our methodology. The larger $r_f$, the farther are neuron's outputs from one other, meaning they exhibit a higher variability. Conversely, smaller values of $r_f$ suggest that the neuron is less relevant, as its outputs are very close in the metric space generated by the unit and can thus be removed. In the case the neuron is removed, to avoid losing the information generated by the neuron, we take the mean plus the standard deviation of the outputs and add it to the bias of the layer. This way, the contribution of the eliminated neuron is implicitly considered in the network's inference process. 

\begin{figure}[h]
	\centering
	\includegraphics[width=0.54\textwidth]{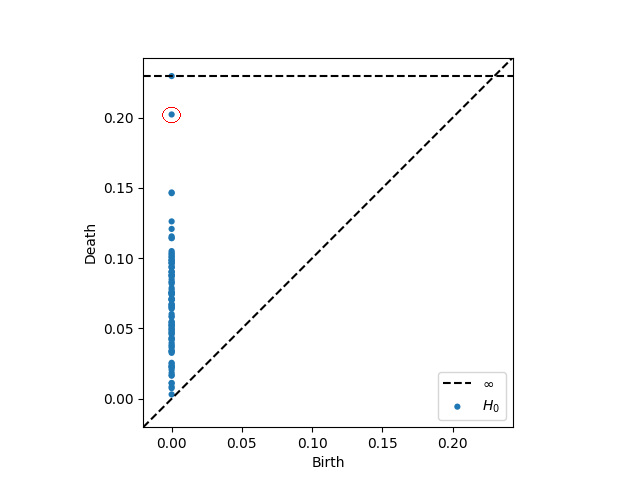}
	\caption{[Diagram: X-axis (Birth Time), Y-axis (Death Time. Persistence)] Here is an example of a Birth-Death persistence diagram. On the X-axis, you find the points where connected components are born. Since we use zero-dimensional persistent homology, all connected components are born at time zero. As the value of $r$ increases (Y-axis), the connected components collapse. Each time two components collapse, a point is represented. The last value below the dashed line corresponds to $r_f$ (circled in red).}
	\label{dgms}
\end{figure}

The analysis of the evolution of connected components as $r$ grows is conducted in the Birth-Death diagram. Since this article focuses on zero-dimensional persistent homology, the points in the diagram align along a line parallel to the Y-axis, since all the points share the abscissa (all connected components are born simultaneously). Therefore, we only consider the values on the Y-axis. $r_f$ can be readily identified in the persistence diagram as the value of $r$ for which all connected components first merge. Figure \ref{dgms} shows an example of persistence diagram in which $r_f = 0.20229639$

\subsubsection{Evaluation of $r_f$ distribution and selection of the important units}
Once we have obtained the value of $r_f$ for each neuron, the analysis the distribution of these values begins. This helps us understand which unit can potentially be suppressed. It leads to a parametric discussion because we can compress the model to a greater or lesser extent depending on how rigorously we evaluate the distribution of the $r_f$ values of all the neurons in the same layer.

The use of statistical metrics facilitates our understanding of how the parameter value is distributed. Essentially, it provides insight into how topologically rich the neuron's contribution is to the overall computation of the network. This is done to maintain a network that is both accurate and efficient. In summary, this approach provides us with explainability regarding the role played by each neuron based on the topological characteristics of its outputs. This clarifies which neurons should be retained in the future simplified model. 

To facilitate the simplification task, we have established three levels of pruning:

\begin{enumerate}
	\item The first, which is the lightest, involves calculating the 30th percentile of the $r_f$ values and retaining neurons with $r_f$ values higher than that percentile.
	\item The second level is slightly more severe, applying the same operation but with the 50th percentile.
	\item The most intense pruning is the third level, where we keep only the neurons with $r_f$ values higher than the 70th percentile.
\end{enumerate}

The more information a layer provides, the higher its chance to remain in the model. This way, we retain the most influential neurons and remove the less relevant ones. After different experiments, we decided to start the simplification process after the second layer. This decision minimally reduces compression capacity but does favor subsequent fine-tuning to meet the requirements of the GLUE benchmark tasks.

\subsubsection{Construction of the compressed model}

The selection of neurons that contribute sufficient information to the original model leads to the construction of the simplified one. This simplified model, which results from the removal in the original model of all the low contribution neurons.

\subsubsection{Evaluation of the compressed model through the GLUE Benchmark}

The evaluation of the simplified model through the GLUE (General Language Understanding Evaluation) benchmark involves assessing the model's performance on a set of diverse natural language understanding tasks. GLUE is a benchmark that consists of multiple downstream NLP tasks, such as text classification, sentence similarity, and question answering. It serves as a standard evaluation suite for assess the generalization and performance of language models.

Here is how the evaluation process generally works:

\begin{itemize}
	\item \textbf{Fine-tuning process:}  The simplified model is fine-tuned on the GLUE benchmark tasks. The tasks are MNLI \cite{williams-etal-2018-broad}, QQP \cite{wang2017bilateral}, QNLI \cite{qnli}, SST-2 \cite{sst2}, CoLA \cite{cola}, STS-B \cite{stsb}, MRPC \cite{mrpc} and RTE \cite{rte}.
	\item \textbf{Model Evaluation:} Use the fine-tuned simplified model to make predictions on the GLUE benchmark tasks. For each task, you will apply the model to the provided test data and obtain predictions.
	\item \textbf{Evaluation Metrics:} Calculate task-specific evaluation metrics for each GLUE task. These metrics can vary depending on the task but often include accuracy, F1 score, or other relevant measures. GLUE provides a standard evaluation script for each task (Table \ref{glue}).
	\item \textbf{Comparison:} Compare the performance of your simplified model to the performance of the original, more complex BERT model and other simplified approaches from the literature. This will give you an indication of how much simplification impacted the model's ability to perform various NLP tasks.
\end{itemize}

The GLUE benchmark provides a standardized way to assess the trade-off between model complexity and task performance. It helps you determine if your simplified model retains sufficient performance on a range of NLP tasks while being computationally more efficient than other BERT models.

\section{Empirical evaluation}

To assess the performance of our methodology, we have designed a thorough empirical procedure that involves a wide range of natural language tasks included in the GLUE benchmark through their datasets (MNLI, QQP, QNLI, SST-2, CoLA, STS-B, MRPC, and RTE, as listed in Table \ref{glue}), and both architectures of the BERT model: BERT Base and BERT Large. This represents a benchmark commonly used in the literature. Additionally, our methodology has been compared with eight state-of-art compression methods for BERT: AUBER \cite{Lee2021-bo}, AE-BERT   \cite{WOS:000890309300033}, ETbLSL    \cite{li-etal-2020-efficient-transformer}, LotteryTicketBert \cite{chen2020lottery}, Michel et al. \cite{michel2019sixteen}, Voita et al  \cite{voita-etal-2019-analyzing} and RPP   \cite{guo2019reweighted}.

\begin{table}[]
	\centering
	\caption{GLUE tasks, train and evaluation sizes and metrics}
	\label{glue}
	\begin{tabular}{|c|c|c|c|}
		\hline
		\textbf{Task} & \textbf{Train} & \textbf{Evaluation} & \textbf{Metric}              \\ \hline
		CoLA          & 10K            & 1K           & Matthew's Correlation        \\ \hline
		SST-2         & 67K            & 872          & Accuracy                     \\ \hline
		MRPC          & 5.8K           & 1K           & F1/Accuracy                  \\ \hline
		STS-B         & 7K             & 1.5K         & Pearson-Spearman Correlation \\ \hline
		QQP           & 400K           & 10K          & F1/Accuracy                  \\ \hline
		MNLI          & 393K           & 20K          & Accuracy                     \\ \hline
		QNLI          & 108K           & 11K          & Accuracy                     \\ \hline
		RTE           & 2.7K           & 0.5K         & Accuracy                     \\ \hline
	\end{tabular}
\end{table}

\subsection{Distribution of $r_f$ and selection of the more informative neurons}

The analysis of the distribution of $r_f$ is crucial for correctly identifying which neurons contribute the most information to a neural network's predictions and which units are dispensable. The premise here is that a neuron with a high associated $r_f$ value implies that its outputs exhibit sufficient variability and distance, making it important when the network computes outputs.

\begin{figure*}[!t]
	\centering
	\includegraphics[width=\textwidth]{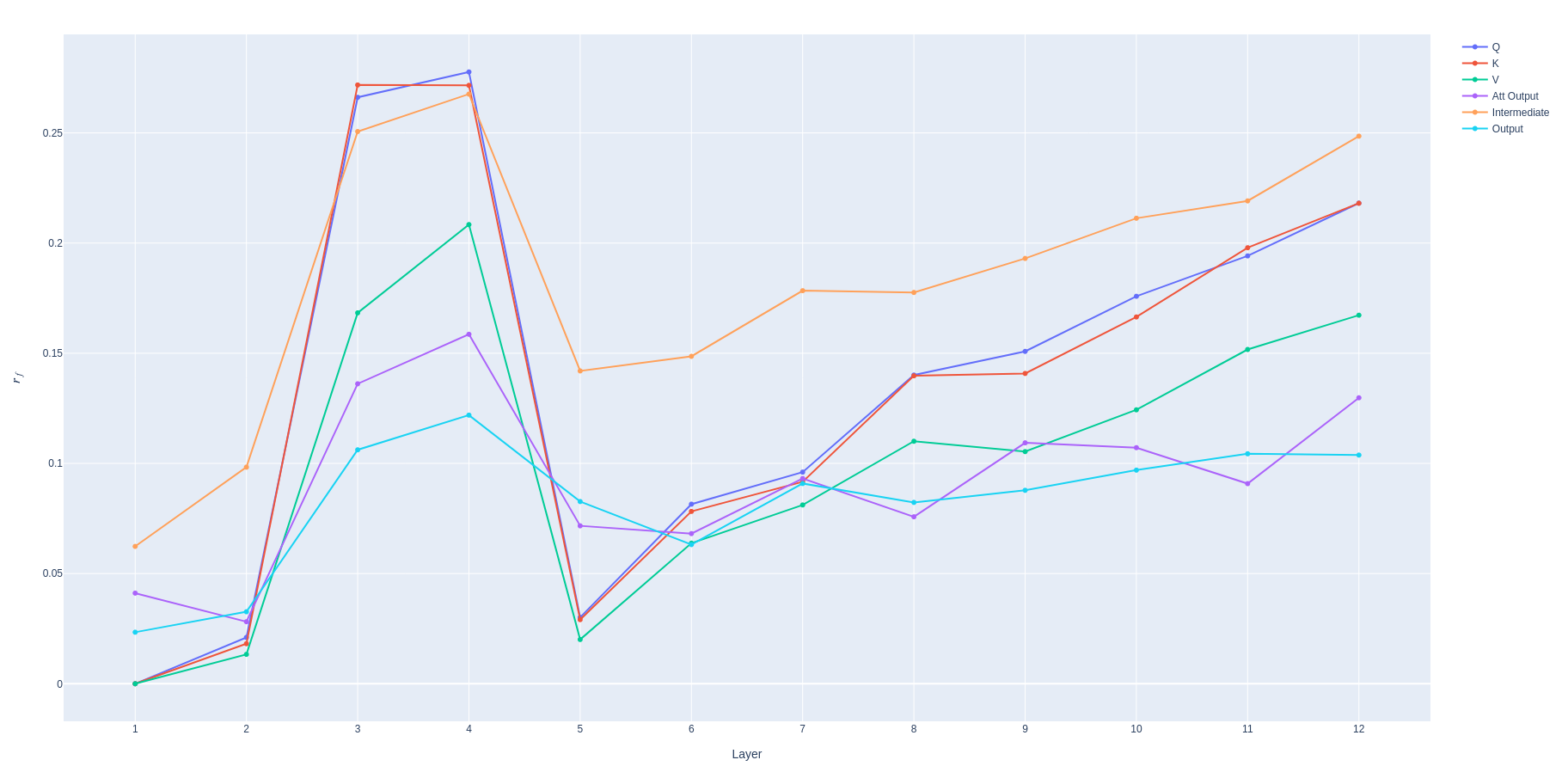}
	\hspace{\fill}
	\includegraphics[width=\textwidth]{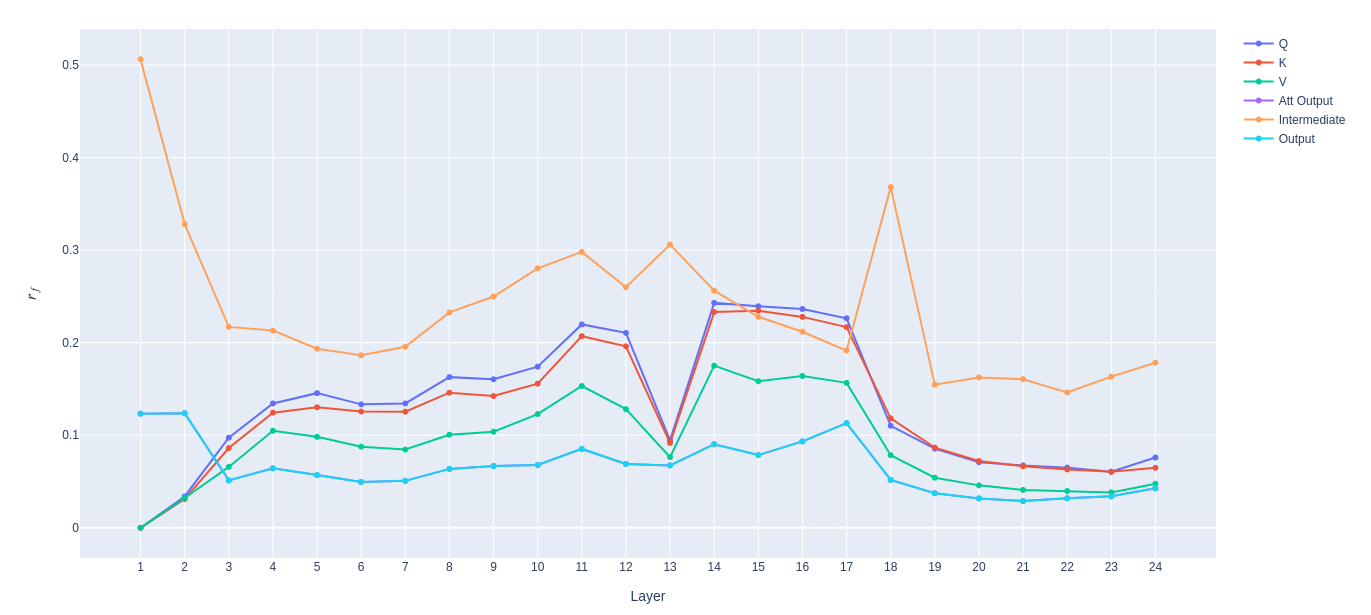}
	\caption{Median per layer of the distribution of $r_f$ for BERT Base (first image) and BERT Large (second image). The components of the \textit{BertLayer} for each layer are shown: Self-Attention Layer (Q, blue; K, red; V, green; Att. Output, purple), Intermediate (orange), and Output (light blue). A higher value of $r_f$ (y-coordinate) indicates a greater contribution of information generated by that component to the network. As can be observed, the flow of information in both networks is different, showing distinct behaviors in each component and layer for BERT Base and Large.}
	\label{0d-ph-images}
\end{figure*}

We noticed that the mean can be heavily influenced by excessively high $r_f$ values in some situations. Therefore, we opted for the median as one of the key metrics to consider. The Figure \ref{0d-ph-images} illustrates the median of the $r_f$ values for each layer and component of BERT Base (left) and BERT Large (right). Before delving into the details of simplification for each architecture, we present some considerations that apply to both BERT Base and BERT Large. Firstly, we observe that the behavior of the Q and K components for each \textit{BertLayer} is very similar, so they have undergone the same pruning process. Additionally, we find that the component that contributes the most information, according to our hypothesis, is the Intermediate component. Lastly, we consistently see that the Attention Output and Output components generate the least information. However, they cannot be simplified because their implementation involves a LayerNorm operation in which the current state of the data and the original input are added together. To assess the importance of this operation, an experiment was performed in which it was removed from the flow. Despite not introducing any simplification, the network lost a significant portion of its predictive capacity, resulting in very poor performance during the fine-tuning process. Therefore, these components cannot be simplified. Now, let's proceed to explain each of these images in detail.

In the case of BERT Base, there is an increasing trend in the contribution of neurons, with outputs becoming more variable and distant as they approach the final layer. This trend experiences an abrupt disruption in layers 3 and 4, where all components exhibit a significant high value for in the median of their $r_f$ values.

In contrast, in the case of BERT Large, the median of the $r_f$ values across the layers does not exhibit a consistent growth pattern. As observed, between layers 3 and 11, all components show increasing values, contributing more information over time. However, there are significant drops in the median of $r_f$ for the Q, K, and V components between layers 12 and 13, as well as between layers 17 and 18. During these same layers, the Intermediate component experiences local maxima. In general, the values are not very high, and the Intermediate component continues to provide a significant amount of information compared to the others, except for layers 15, 16, and 17, where the Q and K components are more informative.

\begin{table}[]
	\centering
	\caption{Levels of pruning applied to each component and layer for the BERT Base architecture. The columns P30, P50, and P70 indicate in which layers light, intermediate, or intense pruning is performed based on their $r_f$ values. Note that the V component tends to contribute less information (most of its layers are heavily pruned), while the Intermediate component provides a significant amount of information (none of its layers are pruned heavily, and most are pruned lightly).}
	\label{comp-bbc}
	\begin{tabular}{|c|c|c|c|}
		\hline
		\textbf{BertLayer Component} & \textbf{P30} & \textbf{P50} & \textbf{P70} \\ \hline
		Q                            & 3, 4, 11, 12    & 8-10         & 5-7          \\ \hline
		K                            & 3, 4, 11, 12    & 8-10         & 5-7          \\ \hline
		V                            & 4            & 3, 11, 12      & 5-10         \\ \hline
		Att. Output                  & -            & -            & -            \\ \hline
		Intermediate                 & 3, 4, 10-12    & 5-9          & -            \\ \hline
		Output                       & -            & -            & -            \\ \hline
	\end{tabular}
\end{table}

\begin{table}[]
	\centering
	\caption{Levels of pruning applied to each component and layer for the BERT Large architecture. The columns P30, P50, and P70 indicate in which layers light, intermediate, or intense pruning is performed based on their $r_f$ values. As in the case of BERT Base, the Intermediate component proves to be relevant for the network's prediction in the BERT Large architecture as well.}
	\label{comp-blc}
	\resizebox{\textwidth}{!}{
	\begin{tabular}{|c|c|c|c|}
		\hline
		\textbf{BertLayer Component} & \textbf{P30}   & \textbf{P50}   & \textbf{P70}      \\ \hline
		Q                            & 11-17          & 4-10, 18       & 3, 13, 19-24      \\ \hline
		K                            & 11-17          & 4-10, 18       & 3, 13, 19-24      \\ \hline
		V                            & -              & 4, 9-12, 14-17 & 3, 5-8, 13, 18-24 \\ \hline
		Att. Output                  & -              & -              & -                 \\ \hline
		Intermediate                 & 3, 4, 8-16, 18 & 5-7, 17, 19-24 & -                 \\ \hline
		Output                       & -              & -              & -                 \\ \hline
	\end{tabular}}
\end{table}

\subsection{Results and analysis on the BERT Base model}
The results for BERT Base are presented in Table \ref{bert-base-table}. In this table, each column corresponds to a task, and each row represents a compression technique. Thus, for each row, the result obtained by each technique is expressed in the metric corresponding to the task (Table \ref{glue}). Results are reported according to the common practices in the literature.

Notice that most state-of-the-art techniques do not perform a complete evaluation on the tasks in the GLUE Benchmark. In our case, we conducted a thorough study of these tasks, reducing the model to $58.47\%$ of the original parameters (from 110M to 64M). Our methodology achieves better results in all the learning tasks compared to state-of-the-art techniques. Even with a significant reduction in the number of parameters, it manages to improve the performance of the original model in most tasks, such as QQP (+20), CoLA (+10.74), STS-B (+1.36), MRPC (+0.73), and RTE (+3.8). This was accomplished with only 40 epochs in the fine-tuning process for each of the tasks.

\subsection{Results and analysis on the BERT Large model}

In contrast to BERT Base, there are far fewer experimental results reported in the state of the art for BERT Large. Once again, our methodology outperforms RPP \cite{guo2019reweighted} in all tasks except for MRPC and RTE, while improving upon the original model in MNLI (0.4/0.33), QNLI (0.45), and CoLA (1.28) (Table \ref{bert-large-table}), again, with only 40 epochs in the fine-tuning process for each of the tasks. In this case, the reduction in the number of parameters is even more significant, resulting in a compressed model with only $52.3\%$ of the parameters (from 340M to 177.2M). This can be explained by the fact that BERT Large is much larger than BERT Base, hence it exhibits greater redundancy.

\begin{table*}[!t]
	\centering
	\caption{Results of compressing BERT Base on GLUE Benchmark tasks. For each metric and task, the higher the value the better (Table \ref{glue}). Best results are highlighted in bold (excluding original BERT)}
	\label{bert-base-table}
	\resizebox{\textwidth}{!}{
	\begin{tabular}{|c|c|c|c|c|c|c|c|c|}
		\hline
		\textbf{Method}   & \textbf{MNLI-(m/mm)} & \textbf{QQP} & \textbf{QNLI} & \textbf{SST-2} & \textbf{CoLA} & \textbf{STS-B} & \textbf{MRPC} & \textbf{RTE} \\ \hline
		Original  \cite{devlin2019bert}        & 84.6/83.4            & 71.2         & 90.5          & 93.5           & 52.1          & 85.8           & 88.9          & 66.4         \\ \hline
		AUBER     \cite{Lee2021-bo}      & -                    & -            & -             & -              & $60.59 \pm 0.73$   & -              & $85.62\pm0.51$   & $65.31\pm1.30$  \\ \hline
		AE-BERT   \cite{WOS:000890309300033}        & -                    & -            & 88.7          & -              & -             & 86.1           & 89.5          & 69.7         \\ \hline
		ETbLSL    \cite{li-etal-2020-efficient-transformer}        & 82.9                 & 90.7         & 88.2          & 89.3           & 52.6          & 84.6           & 88.3          & 63.9         \\ \hline
		LotteryTicketBert \cite{chen2020lottery} & -                    & -            & 88.9          & -              & 53.8          & 88.2           & 84.9          & 66           \\ \hline
		Michel et al. \cite{michel2019sixteen}    & -                    & -            & -             & -              & $58.86\pm0.64$   & -              & $84.22\pm0.33$   & 63.9         \\ \hline
		Voita et al  \cite{voita-etal-2019-analyzing}     & -                    & -            & -             & -              & $55.34\pm0.81$   & -              & $83.92\pm0.71$   & $64.12\pm1.65$  \\ \hline
		QBERT    \cite{shen2019qbert}         & 77.02/76.56          & -            & -             & 84.63          & -             & -              & -             & -            \\ \hline
		OBCE      & \bfseries{83.6/81.6}            & \bfseries{91.23}      & \bfseries{90.03}         & \bfseries{91.8}          & \bfseries{62.96}        & \bfseries{87.16}       & \bfseries{89.63}        & \bfseries{70.2}        \\ \hline
	\end{tabular}}
\end{table*}

\begin{table*}[!t]
	\centering
	\caption{Results of compressing BERT Large on GLUE Benchmark tasks. For each metric and task, the higher the value the better (Table \ref{glue}). Best results are highlighted in bold (excluding original BERT)}
	\label{bert-large-table}
	\resizebox{\textwidth}{!}{
	\begin{tabular}{|c|c|c|c|c|c|c|c|c|}
		\hline
		\textbf{Method} & \textbf{MNLI-(m/mm)} & \textbf{QQP} & \textbf{QNLI} & \textbf{SST-2} & \textbf{CoLA} & \textbf{STS-B} & \textbf{MRPC} & \textbf{RTE} \\ \hline
		Original \cite{devlin2019bert}       & 86.7/85.9            & 72.1         & 92.7          & 94.9           & 60.5          & 86.5           & 89.3          & 70.1         \\ \hline
		RPP   \cite{guo2019reweighted}          & 86.1/85.7            & -            & -             & -              & 61.3          & -              & \bfseries{88.1}        & \bfseries{70.1}      \\ \hline
		OBCE    & \bfseries{87.1/86.23}          & \bfseries{71.9}     & \bfseries{93.15}      & \bfseries{94.7}       & \bfseries{61.78}         & \bfseries{85.2}        & 87.8          & 69.58        \\ \hline
	\end{tabular}}
\end{table*}

\section{Conclusions}

Large Language Models (LLMs) are revolutionizing a number of applications of artificial intelligence, especially in the field of natural language processing. However, due to their complex Transformer-based architecture, LLMs are black-box models with a large number of parameters. In this work, we define OBCE, a methodology aimed at providing explainability to the BERT model, in terms of the role of individual units in the inference and learning process, by using zero-dimensional persistent homology. In particular, OBCE analyzes the topological characteristics of neuron outputs, thereby identifying which neurons contribute more information to the inference process and which ones are dispensable. Even though it can be simply described through a geometric procedure, it has well founded mathematical background based on homology theory. As a result of our successful methodology, we construct simplified versions of the BERT model ($58.47\%$ of parameters for BERT Base, $52.3\%$ for BERT Large) that outperform state-of-the-art techniques, even surpassing the original model's performance for some tasks included in the GLUE Benchmark. This allows us to understand the topological behavior of LLMs like BERT, making them more explainable while maintaining performance and increasing efficiency.

\section{Acknowledgement}

This research has been partially supported by the projects with references TIN2016-81113-R, PID2020-118224RB-100 granted by the Spain’s Ministry of Science and Innovation, and the project with reference P18-TP-5168 granted by Industry Andalusian Council (Consejería
de Transformación Económica, Industria, Conocimiento y Universidades de la Junta de Andalucía), with the cofinance of the European Regional Development Fund (ERDF).

\appendix

\subsection{Homology theory: notation and mathematical background}

Firstly, based on the book \cite{edelcomp} and the survey \cite{10.3389/frai.2021.681108}, in which a thorough explanation of homology theory can be found, we introduce concepts such as affine combination, affine independence, $k-$simplex, face of a $k-$simplex, simplicial complex, chain complexes and the boundary of a $p-$simplex, which are essential for building the theory of homology. Let $v_0, v_1, \dots, v_k$ be vectors in $\mathbb{R}^d$. A point $$x = \sum_{i=0}^{k} \lambda_i v_k$$ is an affine combination of the $v_i$ if $\sum_{i=0}^{k} \lambda_i = 1$. The set of affine combinations constitutes the affine hull. 

\begin{defn}[affinely independent]
	Let $x = \sum \lambda_i v_i, y = \sum \gamma_i v_i$, $x,y \in \mathbb{R}^d$ affine combinations. $x$ and $y$ are affinely independent when $x=y$ if and only if $\lambda_i = \gamma_i \forall i$. In other words, in a plane of dimension $k$ ($k-$plane), $k+1$ points are affinely independent if the $k$ vectors $v_i-v_0$, for $1 \leq i \leq k$, are linearly independent.
\end{defn}

An affine combination $x = \sum \lambda_i v_i$ is a convex combination if $\lambda_i \geq 0$, for $1 \leq i \leq k$. The set of convex combinations is called the convex hull. Now, we introduce the $k-$simplex concept.

\begin{defn}[$k-$simplex]
	A $k-$simplex, $\sigma$, is the convex hull of $k+1$ affinely independent points. Its dimension is $\dim \sigma = k$. 	
\end{defn}

\begin{figure}[h]
	\centering
	\includegraphics[width=0.5\textwidth]{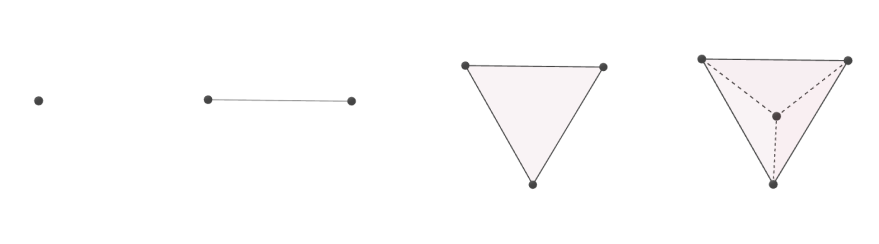}
	\caption{$0-$simplex (vertex), $1-$simplex (edge), $2-$simplex (triangle) and $3-$simplex (tetrahedron)}
	\label{simplexes}
\end{figure}

Note that a $0-$simplex is a vertex, a $1-$simplex is an edge, a $2-$simplex is a triangle and a $3-$simplex is a tetrahedron (Figure \ref{simplexes}). 

A face of a $k-$simplex $\sigma$ is the convex hull of a non-empty subset of $\{v_0, v_1, \dots, v_k\}$. If $\tau$ is a face of $\sigma$, we write $\tau \leq \sigma$.

At this point, we are interested in sets of simplices that are closed under taking faces and that have no improper intersections.

\begin{defn}[simplicial complex]
	A simplicial complex is a finite collection of simplices $K$ such that
	
	\begin{enumerate}
		\item $\tau \in K$  $\forall t \leq \sigma,$ $\sigma \in K$
		\item If $\sigma, \sigma_0 \in K \Rightarrow \sigma \cap \sigma_0$ is empty of a face of both
	\end{enumerate}
\end{defn}

Simplicial complexes ultimately emerge as the intersection of collections of sets. While we have developed this theory considering entirely general simplicial complexes, in practice, we will end up considering the mentioned sets as geometric disks. Thus, the special case of the Vietoris-Rips (VR) complex arises, where the convex sets are disks with radius r. The VR complex will play a fundamental role in our methodology, as it will be crucial for interpreting persistent homology and its application in our approach towards the explainability and simplification of the BERT model.

\begin{defn}[Vietoris-Rips VR complex]
	Let $(P,d)$ a finite metric space. The VR complex of $P$ and $r$ consisting of all subsets of diameter at most $2r$. That is, $\sigma \in \mathbf{R}(r) \Leftrightarrow d(p,q) \leq 2r$, $\forall p,q \in \sigma$.
\end{defn}

After introducing the most basic ideas, we proceed to construct more elaborated concepts that will be part of the definition of the homology group or persistent homology, which are the key tools of this work.

\begin{defn}[Chain complexes]
	Let $K$ be a simplicial complex and $p$ a dimension. A $p-$chain is a formal sum of $p-$simplices, $c = \sum a_i \sigma_i$, where $\sigma_i$ are $p-$simplices and the $a_i$ are modulo $2$ coefficients. 
\end{defn}

Two $p-$chains can be added componentwise. In fact, $p-$chains with the addition operation form the abelian group of $p-$chains $(C_p, +)$.

We define the boundary of a $p-$simplex as the sum of its $(p-1)-$dimensional faces. If $\sigma = [v_0,v_1,\dots, v_p]$ for the simplex defined by the listed vertices, its boundary is 

$$ \partial_p \sigma= \sum_{j=0}^{p} [v_0,v_1,\dots, \tilde{v_j}, \dots, v_p]$$

where $\tilde{v_j}$ is omitted. For a $p-$chain, $c = \sum a_i \sigma_i$, the boundary is $\partial_p c = \sum a_i \partial_p \sigma_i$. Hence, the boundary maps a $p-$chain to a $(p-1)-$chain

$$ \partial_p: C_p(K) \mapsto C_{p-1}(K)$$

Note that $\partial_p(c+c') = \partial_p c + \partial_p c'$, in other words, the boundary is an homomorphism. The chain complex is the sequence of chain groups connected by boundary homomorphisms,

$$ \dots \xrightarrow{\partial_{p+2}} C_{p+1}(K) \xrightarrow{\partial_{p+1}} C_p(K)\xrightarrow{\partial_{p}} C_{p-1}(K) \dots$$ 

The elements of $Z_p(K) = \ker(\partial_p)$ are called $p-$cycles and those of $B_p(K) =$  im$(\partial_{p+1})$ are called $p-$boundaries.

\begin{defn}[Homology group]
	The $p-$th homology group is defined as 
	
	$$ H_p = Z_p / B_p$$
\end{defn}

The $p-$th Betti number is the rank of this group, $\beta_p =$ $rank$ $H_p$.

Finally, we introduce persistent homology, which measures the scale of a topological feature combining geometry and algebra. Consider a simplicial complex, $K$, and a $f:K \mapsto \mathbb{R}$ monotonic, which implies that the sublevel set, $K(a) = f^{-1}(-\infty,a]$ is a subcomplex of $K$ $\forall a \in \mathbb{R}$. Letting $m$ be the number of simplices in $K$, we get $n+1 \leq m+1$ different subcomplexes

$$ \emptyset = K_0 \subseteq K_1 \subseteq \dots \subseteq K_n = K$$

In other words, if $-\infty = a_0 < a_1 < a_2 < \dots < a_n$ are the function values of the simplices in $K$, then $K_i = K(a_i)$. This sequence of complexes is known as the filtration of $f$. For every $i \leq j$, we have an inclusion map from $K_i$ to $K_j$ and therefore an induced homomorphism 

$$f_p^{i,j}:H_p(K_i)\mapsto H_p(K_j)$$

The filtration corresponds to a sequence of homology groups connected by homomorphisms

$$ 0 = H_p(K_0) \mapsto H_p(K_1) \mapsto \dots \mapsto H_p(K_n) = H_p(K)$$

for each dimension $p$. As we go from $K{i-1}$ to $K_i$, new homology classes are gained and other are loosen when they merge with each other. Classes that are born are collected at a given threshold and die after another threshold in groups.

\begin{defn}[Persistent homology]
	The $p-$th persistent homology groups are the images of the homomorphisms induced by inclusion, $H_p^{i,j} = $ im $f_p^{i,j}$, for $0\leq i \leq j \leq n$. The corresponding $p-$the persistent Betti numbers are the rank of these groups.
\end{defn}

We visualize the collection of persistent Betti numbers by drawing points in the extended real plane $\overline{\mathbb{R}^2}$. Letting $\mu_p^{i,j}$ be the number of $p-$dimensional classes born at $K_i$ and dying entering $K_j$, we have

$$\mu_p^{i,j} = (\beta_p^{i,j-1}-\beta_p^{i,j}) - (\beta_p^{i-1,j-1}-\beta_p^{i-1,j})$$

$\forall i<j,$ $\forall p$. Drawing each point $(a_i, a_j)$ with multiplicity $\mu_p^{i,j}$, we get the $p-$th persistence diagram of the filtration.

\end{document}